\documentclass[letterpaper]{article} % DO NOT CHANGE THIS
\usepackage{aaai25}  % DO NOT CHANGE THIS
\usepackage{times}  % DO NOT CHANGE THIS
\usepackage{helvet}  % DO NOT CHANGE THIS
\usepackage{courier}  % DO NOT CHANGE THIS
\usepackage[hyphens]{url}  % DO NOT CHANGE THIS
\usepackage{graphicx} % DO NOT CHANGE THIS
\usepackage{natbib}  % DO NOT CHANGE THIS AND DO NOT ADD ANY OPTIONS TO IT
\usepackage{caption} % DO NOT CHANGE THIS AND DO NOT ADD ANY OPTIONS TO IT
\usepackage{algorithm}
\usepackage{algorithmic}
\usepackage{booktabs} %
\usepackage{newfloat}
\usepackage{listings}
\DeclareCaptionStyle{ruled}{labelfont=normalfont,labelsep=colon,strut=off} % DO NOT CHANGE THIS
\floatstyle{ruled}
\newfloat{listing}{tb}{lst}{}
\floatname{listing}{Listing}
\title{Language and Planning in Robotic Navigation: A Multilingual Evaluation of State-of-the-Art Models}
\author{
    Malak Mansour\equalcontrib,
    Ahmed Aly\equalcontrib,
    Bahey Tharwat\equalcontrib,
    Sarim Hashmi,
    Dong An,
    Ian Reid
}

\affiliations{
    Mohamed bin Zayed University of Artificial Intelligence\\
    Masdar City, Abu Dhabi, UAE\\
    \texttt{\{malak.mansour, ahmed.aly, bahey.tharwat, sarim.hashmi, dong.an, ian.reid\}@mbzuai.ac.ae}
}

\usepackage{bibentry}

\begin{document}

\maketitle

% TLDR
% This study explores the integration of Arabic language in Vision-and-Language Navigation (VLN) using the NavGPT framework. By evaluating multilingual Small Language Models (SLMs) and the Arabic-centric LLM, Jais, on navigation tasks with English and Arabic instructions, we reveal that while some models excel in planning, others struggle with reasoning in Arabic due to inherent limitations. Our findings emphasize the need to enhance language models' reasoning and planning capabilities, especially for Arabic, to unlock their potential for real-world applications.

% We evaluate multilingual and Arabic-centric language models for vision-and-language navigation, highlighting their strengths and limitations in reasoning and planning with Arabic instructions.

\begin{abstract}
% DESCRIPTION: Write a concise abstract (one paragraph) of your project. Highlight the problem, your approach, key findings, and the significance of your work. Also give link of your Github repository. Last line of the abstract should be like this:\\
Large Language Models (LLMs) such as GPT-4 exhibit strong reasoning and planning capabilities. This work introduces the first study on Arabic-language integration in Vision-and-Language Navigation (VLN) in Robotics. We evaluate multilingual Small Language Models (SLMs) such as GPT-4o mini, Llama 3 8B, Phi-3 medium 14B, and the Arabic-centric Jais LLM using the NavGPT framework on the R2R dataset, augmented with Arabic translations.

Our experiments show that the framework supports high-level planning in both English and Arabic. However, some models struggled in Arabic due to parsing and capability limitations. This highlights the need to improve planning in LMs and reveals the potential of Arabic-centric models in real-world robotics.

Code is available at \\
\url{https://github.com/Malak-Mansour/Language-In-RobNav}
\newline
\textit{\textbf{Keywords} Vision-and-Language Navigation (VLN), Small Language Models (SLMs), Large Language Models (LLMs), Jais, NavGPT, Reasoning, Robotics}

\end{abstract}

\section{Introduction} 
\label{s:intro}
% DESCRIPTION: Clearly define the problem or objective of your project. Explain the motivation behind your choice of topic. Provide a brief overview of the methods and techniques you used. Failure to do so will result in marks deduction. (must include problem motivation) 

With the rise of AI-driven robotics in smart cities, multilingual interaction systems have become increasingly crucial, particularly in the Middle East and North Africa (MENA) region, where investment in autonomous systems is growing rapidly. However, the scarcity of models trained on Arabic data presents a significant barrier to their deployment. Despite Arabic’s importance, spoken by over 400 million people \cite{ArabicMMLU}, its under-representation in Vision-and-Language Navigation (VLN) research limits the effectiveness of autonomous systems in Arabic-speaking regions. This research addresses these gaps by examining how Arabic and English inputs influence the planning and reasoning capabilities of Language Models (LMs) in robotic VLN tasks. The goal is to contribute to developing more inclusive autonomous systems that address the linguistic and cultural diversity of the Arabic-speaking world by seamlessly understanding and executing navigation instructions in Arabic.

Small Language Models (SLMs) are more efficient and cost-effective than Large Language Models (LLMs) due to their smaller size and reduced computational requirements, making them suitable for deployment on resource-constrained devices \cite{SLMs}. Their smaller footprint enables faster response times and easier integration into existing systems. Despite these advantages, SLMs have received comparatively less attention in research, presenting an opportunity to explore their potential for optimization, efficiency, and domain-specific applications in both language understanding and high-level planning for navigation tasks.

In this work, we focus on SLMs rather than Vision-Language Models (VLMs) due to SLMs' superior ability to process and reason about complex linguistic structures across multiple languages \cite{SLMs}. As noted by \cite{visionlanguagemodelsblind}, VLMs often face performance limitations in real-world navigation tasks, particularly because their approach to extracting visual features tends to neglect instruction prompts, thereby reducing adaptability and planning efficiency. Moreover, excluding vision from the reasoning process mitigates potential simulation-to-reality gaps during robot deployment, ensuring that systems are more robust in their ability to execute plans effectively in diverse real-world scenarios.

\begin{figure*}[hbt!]
    \centering
    \includegraphics[width=\linewidth]{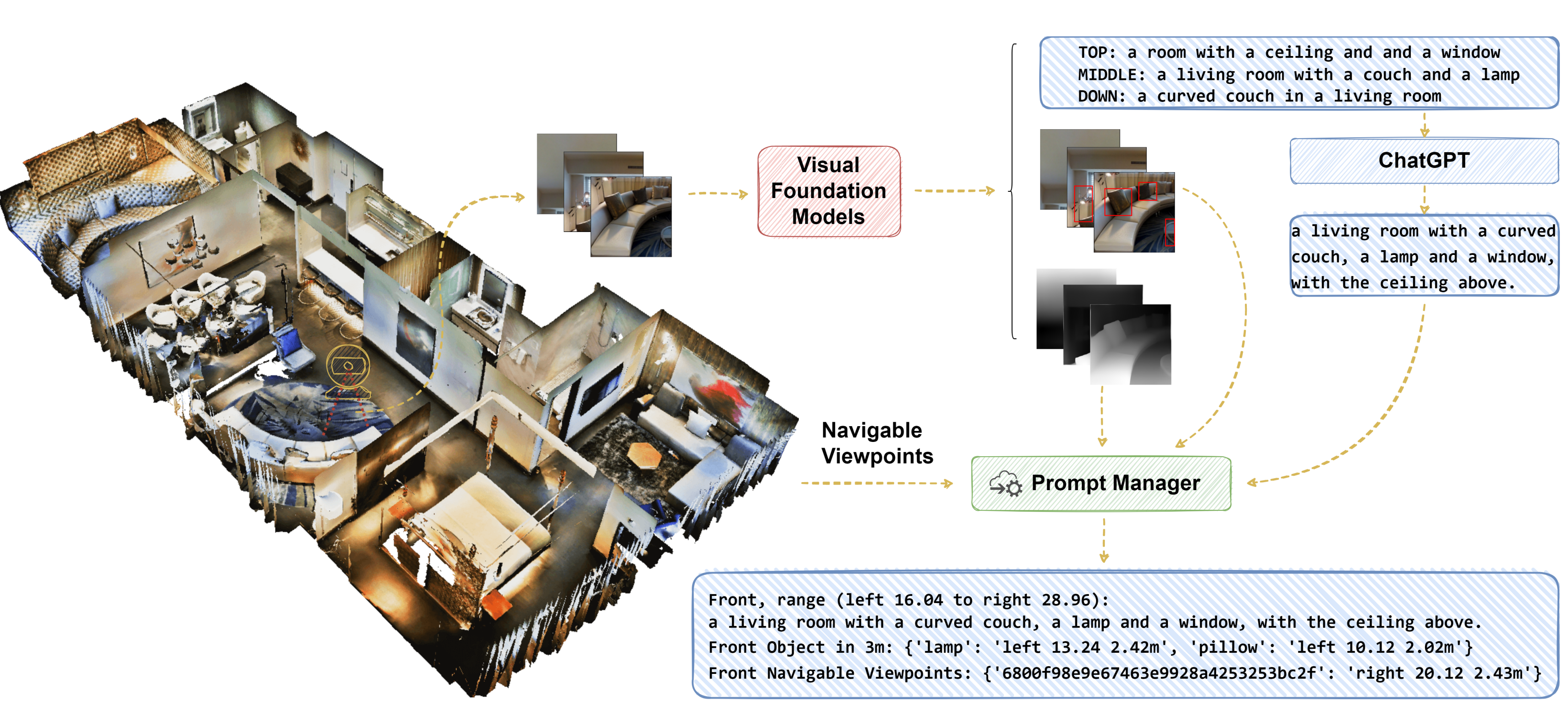}
    \caption{NavGPT methodology diagram \cite{NavGPT}}
    \label{fig:Navgpt_flowchart}
\end{figure*}

Our evaluation involves inferencing and comparing state-of-the-art multilingual SLMs and Core42’s Arabic-centered LLM, Jais 30B \cite{Jais}, in the NavGPT framework. The SLMs are OpenAI’s GPT-4o mini \cite{gpt4omini}, Meta’s Llama 3 8B \cite{llama}, and Microsoft’s Phi-3 medium 14B \cite{phi}. Using the Room-to-Room (R2R)-VLN dataset \cite{R2R}, which provides English-language navigation instructions, we augment the data with Arabic translations generated using the Groq API for comparative analysis. The models are evaluated within the NavGPT framework, which operates in a zero-shot manner to predict sequential actions based on textual descriptions of visual observations, navigation history, and navigable viewpoints, enabling a systematic assessment of their planning and reasoning capabilities, as illustrated in Figure \ref{fig:Navgpt_flowchart} \cite{NavGPT}.

\subsection*{Contributions}
The contributions of this work are as follows:
% DESCRIPTION: Write contributions of your project in bulleted list.
\begin{itemize}
    \item Developed the first-ever framework incorporating Arabic instructions in the VLN problem to evaluate the planning capabilities of multilingual SLMs (GPT-4o mini, Llama 3 8B, Phi-3 medium 14B) and the Jais 30B LLM on VLN tasks using both Arabic and English instructions.
    \item Augmented the R2R-VLN dataset with Arabic translations via the Groq API, creating a bilingual dataset for VLN research to address linguistic and planning challenges in diverse environments.

    \item Conducted an in-depth analysis of the impact of instruction language on the reasoning and planning capabilities of LMs.
    % \item Designed a robust experimental setup using NavGPT \cite{NavGPT} to evaluate navigation performance and mitigate the risk of model hallucination.
    \item Identified key insights into language-specific limitations and strengths in multilingual SLMs and the Jais LLM, providing a foundation for developing autonomous systems capable of robust planning and execution in Arabic-speaking regions.
    % \item Proposed a future deployment strategy for an assistive robotic navigation system that processes Arabic voice instructions, aiming to enhance accessibility for visually impaired users in real-world environments.
\end{itemize}

\section{Related Works}
\label{s:related-work}

%DESCRIPTION: Summarize relevant research and prior work in the field. Explain how your project fits into the existing body of knowledge. Throughout the project report, cite the papers where necessary. Failure to do so will result in marks deduction. 

\subsection{Classical planners}

Recent classical planning systems in robotics commonly employ Planning Domain Description Language (PDDL) or Answer Set Programming (ASP) as the underlying action language for planners \cite{Ghallab_Nau_Traverso_2016, jiang2019taskplanningroboticsempirical}. PDDL is a formal language designed for expressing planning problems and domains. It provides a structured way to define the initial state, goal state, and possible actions, enabling planners to generate valid sequences of actions to transition from the initial state to the goal state. It has been widely used due to its expressiveness and ability to handle various planning paradigms. ASP, on the other hand, is a declarative programming approach rooted in logic programming and non-monotonic reasoning. It allows for the encoding of complex planning tasks as sets of logical rules and constraints, making it well-suited for tasks requiring reasoning over discrete domains and handling combinatorial complexity.

These methods are strong in ensuring soundness, completeness, and efficiency in structured environments where tasks and states are well-defined. However, they struggle in open-world or dynamic scenarios, where predefined actions and states cannot always accommodate the unpredictability of the real world as they do not consider scene understanding. On the other hand, VLN models, which combine scene understanding and language instruction, have gained attraction in robotics \cite{NavGPT}. These models allow robots to interpret and execute actions sequentially in every state based on visual inputs and natural language instructions, making them more adaptable to dynamic environments.

\subsection{VLN Task}
VLN tasks use natural language instructions to guide agents through diverse environments, testing the reasoning and comprehension capabilities of LMs \cite{survey}. These tasks typically evaluate agents in indoor settings, such as navigating through rooms in a building \cite{R2R}, or outdoor scenarios, such as traversing open landscapes or urban areas \cite{Openstreetmap}. Models must interpret navigation commands, align them with visual context, and generate appropriate actions in these varied environments.

Despite their promise, VLN tasks pose significant challenges. Ambiguities in language instructions, such as vague descriptions or inconsistent phrasing, can hinder accurate alignment with visual cues \cite{R2R}. Sparse or incomplete visual data, such as in dim indoor settings or cluttered outdoor scenes, complicate navigation \cite{visionlanguagemodelsblind}. These challenges are exacerbated when operating in multilingual contexts, where instructions may carry linguistic nuances that models struggle to capture or disambiguate. Addressing these limitations is crucial for improving real-world deployment of VLN systems \cite{RxR,unseenenvironments}.

\subsection{Language Models in Navigation}
LMs play a pivotal role in VLN tasks by interpreting instructions and guiding agents through diverse environments. SLMs, LLMs, and VLMs each have distinct capabilities. SLMs are lightweight and efficient, enabling real-time processing in resource-constrained settings \cite{SLMs}. Their streamlined architecture ensures practical deployment, though they may lack the extensive training and generalization capabilities of LLMs. By contrast, LLMs excel in multilingual reasoning and zero-shot tasks due to their vast training on diverse datasets \cite{NavGPT, NavGPT2}. However, their computational demands and reliance on text-only reasoning limit their suitability for scenarios requiring multimodal integration \cite{Vilbert}.

VLMs specialize in combining visual and textual inputs, excelling in spatial reasoning tasks where strong visual grounding is critical. Despite this, VLMs often underperform in linguistically complex or multilingual scenarios, as their architectures prioritize visual processing over comprehensive language understanding \cite{visionlanguagemodelsblind, ArabicMMLU}. This study focuses on SLMs' ability to handle linguistic challenges like Arabic’s morphological richness and syntactic complexity. This can enhance model robustness through exposure to diverse linguistic structures while ensuring efficient deployment in real-world applications, particularly in resource-constrained environments common in the MENA region.

\subsection{Multilingual Challenges and Arabic-Specific Context}
While most VLN studies focus on English-language instructions, they overlook the complexities and opportunities of non-English languages \cite{survay_foundation}. This work explores multilingual capabilities, particularly Arabic. Arabic, a widely spoken but underrepresented language in VLN, introduces unique challenges due to its morphological richness, syntactic complexity, and right-to-left script \cite{Arabic_AI}. These linguistic features make Natural Language Processing (NLP) tasks, including VLN, more demanding. Furthermore, the scarcity of high-quality Arabic datasets exacerbates these challenges, limiting the performance and adaptability of existing models \cite{RxR}.

Arabic-centric models like Jais \cite{Jais} provide a foundation for addressing these gaps by leveraging training data tailored to Arabic. However, many VLN datasets, such as R2R \cite{R2R}, remain focused on English, requiring augmentation or translation to support multilingual research. This study tackles these issues by analyzing how Arabic-language instructions impact reasoning in state-of-the-art models and identifying areas where existing architectures fall short in supporting Arabic tasks.

\subsection{Datasets and Simulation Environments}
VLN research relies heavily on paired visual and linguistic datasets. The R2R dataset \cite{R2R}, with its English-language instructions for navigating photo-realistic environments, serves as a standard benchmark for evaluating VLN models’ spatial reasoning and language comprehension. Simulation environments, such as Matterport3D \cite{Matterport3D}, are crucial for testing models in realistic indoor settings, providing diverse and visually rich scenarios for evaluating navigation performance. Building on R2R, the RxR dataset \cite{RxR} introduces multilingual instructions, including languages like Hindi, but excludes Arabic, which facilitates research on cross-linguistic adaptability.

Currently, Arabic navigation datasets are scarce. OpenStreetMap \cite{Openstreetmap}, a multilingual outdoor navigation dataset, includes limited non-English instructions, though Arabic coverage remains minimal. Given these limitations, our study leverages a translated R2R dataset to compare English and Arabic reasoning, highlighting language effects on navigation accuracy and providing groundwork for broader multilingual VLN research.

\section{Problem Statement}
\label{s:prob-statement}
%DESCRIPTION: In your project report's problem statement, succinctly state the issue you are addressing, providing context, significance, and specific details.
The MENA region's growing reliance on autonomous systems emphasizes the need for multilingual VLN models capable of understanding Arabic, a language spoken by over 400 million people \cite{ArabicMMLU}. Arabic’s morphological richness and syntactic complexity present unique challenges for NLP \cite{Arabic_AI}, making it a valuable test case for evaluating language model reasoning. Despite this importance, existing VLN datasets, such as R2R, are predominantly in English \cite{R2R}, leaving a critical gap in resources for Arabic-language tasks.

LLMs like GPT \cite{GPT4} and Jais \cite{Jais} have demonstrated strong multilingual reasoning capabilities, but their performance in Arabic navigation contexts remains under-explored. SLMs are cost-effective and practical for deployment on resource-constrained devices, such as for real-time navigation; however, their potential remains underrepresented in VLN research. Few studies have compared SLMs to LLMs in the context of multilingual navigation, highlighting a gap in the literature \cite{gpt4omini}. Addressing these gaps, this study explores how Arabic and English instructions affect reasoning capabilities in state-of-the-art SLMs and LLMs, aiming to improve inclusivity and adaptability in robotic navigation systems.

\section{Methodology}
\label{s:methodology}

%DESCRIPTION: Clearly explain the method that you are using to solve the problem. Add \textbf{diagrams} and necessary \textbf{mathematical formulation} to make the reader understand the method.
% Briefly explain the deep learning techniques and models you used in detail. Justify your choice of methods and algorithms.

\begin{figure}
    \centering
    \includegraphics[width=1\linewidth]{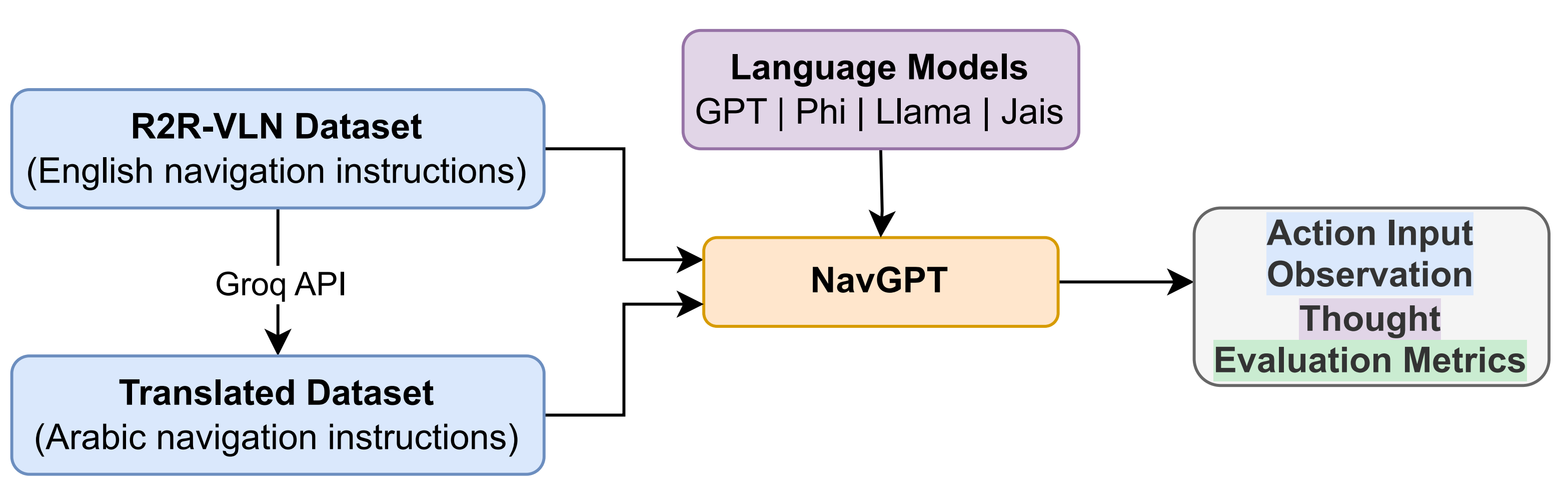}
    \caption{Evaluation pipeline}
    \label{fig:Project_flowchart}
\end{figure}

In this section, we outline the problem formulation, method for translating the English R2R dataset to Arabic, and how NavGPT, a state-of-the-art LM-based navigation system, is adapted to process Arabic instructions. Our approach encompasses dataset translation, model setup, inference execution, and performance evaluation laid out in the pipeline in Figure \ref{fig:Project_flowchart}.

\subsection{VLN Problem Formulation}

NavGPT addresses the VLN problem by framing it as follows \cite{NavGPT}. Given a natural language instruction \( W \), represented as a sequence of words {\( w_1, w_2, w_3, \dots, w_{n} \)}, the agent retrieves an observation \( O \) at each step \(s_t\) by interpreting its current location through a simulator. This observation consists of \( N\) alternative viewpoints, representing the surrounding environment of the agent in varying angles.

Each unique view observation is denoted as \( o_{i} \, (i \leq N) \), with its corresponding angle direction represented as \( a_{i} \, (i \leq N) \). Consequently, the observation at step $t$ can be expressed as:
\[
O_t = \left( \left( o_1, a_1 \right), \left( o_2, a_2 \right), \dots, \left( o_N, a_N \right) \right)
\]
During navigation, the agent's action space is restricted to the navigation graph \( G \), where G is a predefined graph of navigable viewpoints. At each step, the agent selects the next action from the \( M = |C_{t+1}| \) set of navigable viewpoints, \( C_{t+1} \). This selection is guided by aligning the current observation \( O_{C_t} \) with the provided instruction \( W \). The agent predicts the next action by identifying the relative angle \( a_{C_i} \) from \( O_{C_t} \), executes this action through interaction with the simulator, and transitions from the current state \( s_t = (v_t, \theta_t, \phi_t) \) to the next state \( s_{t+1} = (v_{t+1}, \theta_{t+1}, \phi_{t+1}) \), where \( v \), \( \theta \), and \( \phi \) represent the agent's current viewpoint, heading, and elevation angle, respectively. 

To support navigation, the agent maintains a history of its previous states \( h_t \) and updates the conditional transition probability between states as follows:
\[
S_t = T \left( s_{t+1} | a_{C_i}, s_t, h_t \right)
\]
where \( T \) represents the conditional transition probability distribution.

In summary, the agent learns a policy \( \pi \) parametrized by \( \Theta \) that relies on the instruction \( W \) and the current observation \( O_{C_t} \), expressed as:
\[
\pi(a_t | W, O_t, O_{C_t}, S_t; \Theta)
\]
This study conducts the VLN task in a zero-shot setting, where \( \Theta \) is not trained using VLN-specific datasets but is instead derived from the language corpus used to train the LMs.

\subsection{Dataset Translation}
To convert the English R2R dataset to Arabic, we used the Groq API, specifically the Llama-3.2-90B-text-preview model. We then developed a simple prompt-based code to translate the English instructions, objects list, and observations in the R2R dataset to Arabic, ensuring that the translated dataset maintained the same format as the original English version for seamless compatibility with NavGPT. This alignment facilitates using both the original and translated datasets in comparative experiments.

\subsection{Incorporating LMs}
To explore different language models with NavGPT, we utilized LMs deployed on Azure for inference, enabling a comparative analysis. Following NavGPT’s pipeline shown in Figure \ref{fig:Navgpt_flowchart}, we configured the SLMs (GPT-4o mini, Llama, Phi-3) and Jais 30B LLM, and tested them on the English and Arabic datasets to facilitate a direct comparison of their performances.

The selected SLMs were chosen for their diverse architecture sizes, multilingual capabilities, and unique approaches to processing input. GPT-4o mini, with its compact architecture, demonstrates efficiency in understanding and reasoning across languages by leveraging large-scale multilingual training \cite{gpt4omini}. Llama 3 8B excels in instruction-following tasks, combining a medium-sized model with strong contextual understanding \cite{llama}. Phi-3 Medium 14B balances scalability and reasoning power, enabling nuanced task-specific performance \cite{phi}. Jais 30B, optimized for Arabic, enhances linguistic diversity by deeply integrating Arabic-specific datasets, ensuring accurate comprehension and generation \cite{Jais}. These models allow for analyzing the effects of multilingual input on reasoning, focusing on whether instructions are directly reasoned upon or internally translated.

% comparing model sizes: https://techcrunch.com/2024/07/18/openai-unveils-gpt-4o-mini-a-small-ai-model-powering-chatgpt/

% https://www.unite.ai/gpt-4o-mini-unveiled-a-cost-effective-high-performance-alternative-to-claude-haiku-gemini-flash-and-gpt-3-5-turbo/

\subsection{Inference with NavGPT}
The NavGPT framework integrates natural language instructions and visual observations for autonomous navigation. Instructions are processed alongside environmental data using Visual Foundation Models, which extract key features from the current viewpoint. A Prompt Manager formats this information into structured inputs for an LM, which reasons over the trajectory to decide the next action or stop. A history buffer tracks previous states to ensure consistent decision-making. This pipeline facilitates robust multimodal navigation with real-time reasoning capabilities tailored to user instructions.

We ran NavGPT in inference mode using the SLMs and Jais for a subset of the data. We assessed the configured LMs' effectiveness on English and Arabic instructions and evaluated their performance to see if language affects reasoning in navigation. We used 100 sample trajectories from the \textit{val unseen} dataset in the R2R dataset. Each trajectory output includes:
\begin{itemize}
    \item \textbf{Action Input}: The upcoming trajectory ID
    \item \textbf{Observation}: Textual descriptions of the environment at each location
    \item \textbf{Thought}: The robot’s thoughts, reasoning, and planning about reaching the target location and identifying obstacles
    % \item \textbf{Evaluation Metrics}: action steps, steps, lengths, nav error, oracle error, SR (success rate), oracle SR, SPL (success rate weighted by length), nDTW, SDTW, CLS
    \item \textbf{Evaluation Metrics}: Action steps, total steps, path lengths, navigation error, oracle error, success rate (SR), oracle success rate (oracle SR), success weighted by path length (SPL), normalized dynamic time warping (nDTW), success weighted by dynamic time warping (SDTW), and coverage length score (CLS)
\end{itemize}

Through these steps, our methodology combines translation, inference, and evaluation, providing a structured approach to deploying NavGPT for Arabic-language navigation tasks. This approach ultimately allows us to measure the model’s efficacy across languages and guide improvements for multilingual robotic navigation.

\section{Experimental Setup}
\label{s:experimental-setup}
%DESCRIPTION: Describe the hardware and software environment used for your experiments. Detail the hyperparameters and settings for your models. Also give details of the baseline method(s) used in the project to comapre results.

The experiments were conducted using a combination of local hardware and cloud-based APIs. The local setup included machines with NVIDIA Quadro 6000 GPUs, each with 24 GB of memory, primarily for dataset preparation and evaluation tasks. The models evaluated—GPT-4o mini \cite{gpt4omini}, Llama 3 8B \cite{llama}, Phi-3 Medium 14B \cite{phi}, and Jais 30B \cite{Jais}—were hosted on Azure’s serverless platform and accessed via APIs in the same configuration. This setup ensured consistency in model performance while leveraging Azure’s scalability. For dataset augmentation, the Groq API was used to generate Arabic translations of English instructions in the R2R dataset.

This study evaluates the zero-shot reasoning capabilities of pre-trained language models, focusing on their ability to handle navigation tasks in both English and Arabic. No training or fine-tuning was performed. Input instructions and navigation trajectories were fed directly to the models via APIs without modifications to the underlying model architecture. To ensure outputs were in the correct format, different prompts were used depending on the model, aligning responses with the required structure for evaluation. This consistent configuration allowed for a controlled and fair comparison across all models.

\subsection{Dataset}
The evaluation utilized the R2R dataset alongside its Arabic-augmented counterpart. Arabic translations were generated using the Groq API, maintaining alignment with the original English instructions. A total of 100 navigation trajectories were evaluated in each language. The same 100 trajectories from the English dataset were used in the augmented Arabic dataset to ensure consistency. This framework allowed for a direct comparison of language-specific model reasoning and navigation capabilities.

\subsection{Evaulation Metrics}
%DESCRIPTION: Explain the evaluation metric(s) used in the project.
\subsubsection{Quantitative assessment}
We compared the models with each other using the following standard evaluation metrics for VLN tasks \cite{R2R}:
\begin{itemize}
    \item \textbf{Trajectory Length (TL)}: the average distance traveled by the agent during navigation
    \item \textbf{Navigation Error (NE)}: the mean distance between the agent's final location and the target location
    \item \textbf{Success Rate (SR)}: measures the percentage of completed trajectories where the robot reaches its goal
    \item \textbf{Oracle Success Rate (OSR)}: evaluates whether the agent was on the right path even if it didn’t stop at the exact target location
    \item \textbf{Success weighted by Path Length (SPL)}: considers the length of the trajectory relative to the shortest path
\end{itemize}

Quantitative results are shown in Table \ref{tab:performance_metrics}.

\subsubsection{Qualitative Assessment}

We also conducted a qualitative assessment to examine the models' performance, focusing on their reasoning and decision-making processes. This evaluation highlights subjective observations that help identify weaknesses in the models' reasoning and planning capabilities. Specifically, we assessed the following aspects:
\begin{itemize} 
    \item \textbf{Reasoning}: Analyzed how effectively the models interpreted navigation instructions, integrated visual observations, and decomposed complex instructions into actionable sub-goals.
    \item \textbf{Spatial Awareness}: Evaluated the models' ability to comprehend their current environment, maintain navigation history, and use this information to make accurate decisions.
    \item \textbf{Planning}: Investigated how the models structured sequential steps, anticipated future states, and adjusted their plans dynamically based on the environment.
\end{itemize}

\section{Results and Discussion}
\label{s:results-discussion}
%DESCRIPTION: Present the results of your experiments, including quantitative and qualitative findings. Use tables and visualizations to support your claims. Give comments on the results and explain the findings.

This study explored the performance of various language models in reasoning and understanding complex navigation instructions in English and Arabic. The models tested included GPT-4o mini, Llama 3, Phi-3, and Jais, with significant variations in their ability to parse and execute instructions. Furthermore, the qualitative assessment provides textual examples of some of the model outputs, including failed cases. Based on the models' performance, we categorized them into three groups: Working, Partially Working, and Not Working. We show that the relative limitations of LMs depend on the input language and planning aspects, which affect their response to the navigation instructions.

\subsection{Quantitative Assessment}
 % Table \ref{tab:model_comparison} summarizes the qualitative performance of the models, listing their pros and cons, while 
Table \ref{tab:performance_metrics} presents the aggregated results of quantitative evaluation metrics. During inferencing, NavGPT outputs thoughts that demonstrate its reasoning process as it navigates the environment. We prompted the model to output its thoughts in Arabic whenever we were inferencing with the Arabic-translated dataset as input, creating a monolingual Arabic context. Only with GPT-4o mini did we mix input languages, combining Arabic datasets and English thoughts, to observe how this robust model performs.

We evaluated the models on 100 trajectories. However, some of the language models frequently ran out of context window when reasoning through more complex long instructions. This occurred primarily with smaller or less robust models, such as Phi and Jais, as reflected in their lower number of successful predictions out of 100 trajectories (Table \ref{tab:performance_metrics}).

NavGPT relies on structured prompts to perform optimally in navigation tasks. Specifically, the input should include a well-defined task description, such as goal location and intermediate waypoints. When these structured inputs are missing or incomplete, the model often struggles to generate accurate predictions, as shown in Figures \ref{fig:Errors} and \ref{fig:Jais_Error}. These errors typically happen when the model cannot fully reason over the provided instructions, resulting in its failure to output the necessary information in the required format for continued navigation.

\begin{itemize}

\item \textbf{Working:} 

    \begin{enumerate}
    \item \textbf{GPT-4o mini:} This model is trained on a large multilingual dataset, eliminating the need for explicit translations of non-English inputs. As a result, it successfully handles both English and Arabic datasets. Its performance metrics for monolingual English and Arabic scenarios were comparable, achieving the highest values for Trajectory Length (TL), Success Rate (SR), Oracle Success Rate (OSR), and Success weighted by Path Length (SPL), while maintaining the lowest Navigation Error (NE) compared to other models, showcasing its robustness. However, the mixed scenario of Arabic data with English reasoning experienced slightly higher navigation error, lower SR and SPL, and a marginally higher OSR (37.00) than the pure Arabic scenario (36.00). This discrepancy could be attributed to a misalignment between the Arabic dataset and English reasoning. Nevertheless, GPT-4o mini (SR=21) outperformed the next best model, Phi-3 (SR=7), by approximately three times and Llama 3 (SR=4) by approximately five times.

    \item \textbf{Llama 3 8B:} Llama 3 exhibited reasonable reasoning and planning capabilities when processing both English and Arabic instructions. However, it fell short of GPT-4o mini’s performance metrics, likely due to its smaller multilingual training dataset and less optimization for diverse linguistic tasks. Its SR (4 for English and 3.12 for Arabic) indicates limited success in executing goal-oriented tasks. Despite these challenges, Llama 3’s ability to handle Arabic instructions suggests it holds promise for future development in multilingual reasoning. Even with its smaller size (8B), Llama 3 successfully completed nearly all 100 trajectories with both datasets, demonstrating its robust capabilities and large context window.
    \end{enumerate}

\item \textbf{Partially Working (Phi-3 medium): } 

    Phi-3 medium demonstrated competitive performance in processing English instructions but faced challenges due to a smaller number of successful predictions (41/100) and parsing issues with the \textit{viewpoint ID}. These issues were likely caused by the model's strict format requirements for input alignment, highlighting its lack of robust natural language understanding. This limitation led to incorrect outputs, as detailed in the qualitative assessment. 

    For Arabic tasks, Phi-3 failed entirely, which can be attributed to its non-multilingual nature and insufficient exposure to Arabic language data during training. Consequently, it was unable to process or generate meaningful outputs in Arabic. Moreover, the model only evaluated 18 out of 100 trajectories, revealing its limited robustness in handling complex tasks in Arabic.

\item \textbf{Not Working (Jais): } 

    % Jais 30B, the only Arabic-centric LLM in this experiment, surprisingly exhibited poor reasoning capabilities in both Arabic and English. It performed the worst across all critical metrics. Despite expectations that it would perform well due to its Arabic focus, Jais 30B struggled with reasoning, potentially due to issues in its fine-tuning techniques or system prompting, which hindered its ability to adapt to the navigation task requirements. Nevertheless, its large size allowed it to complete 82 out of 100 trajectories.
    Jais 30B, the only Arabic-centric LLM in this experiment, surprisingly exhibited poor reasoning capabilities in both Arabic and English, performing the worst across all critical metrics. Although it was expected to perform well due to its Arabic focus, Jais 30B struggled with reasoning in the context of navigation tasks. This poor performance may be attributed to its initial training by Core42, which was not specifically optimized for navigation-related tasks. However, despite these limitations, Jais 30B's large size allowed it to complete 82 out of 100 trajectories.

\end{itemize}

This analysis demonstrates that the underlying architecture and multilingual training of the LMs are more influential than the input language in determining performance. This conclusion is supported by the smaller performance variation observed within robust LMs like GPT-4o mini across English and Arabic datasets compared to the larger differences between models. Therefore, selecting high-capacity, multilingual LMs that can generalize across languages can mitigate the need for language-specific fine-tuning for navigation tasks.

% Key issues included parsing errors caused by non-standard formats and poor reasoning capabilities, especially for Arabic instructions. These findings highlight the limitations of smaller models in complex planning tasks.

% Following the successful performance of GPT-4o mini, The mixed dataset and instructions experiment with GPT-4o mini performed worse than the pure lingual English or Arabic. It reported a lower trajectory length, higher error, lower SR and SPL, but slightly higher OSR (37.00) than the pure Arabic (36.00) as shown in Table \ref{tab:performance_metrics}. This could be attributed to the back-and-forth translations between Arabic and English, accumulating the translation errors forward.

\begin{table}[ht]
\centering
\small % Set font size to 9-10pt Roman
\caption{Quantitative Analysis of the LMs with English and Arabic Datasets}
\setlength{\tabcolsep}{1mm} % Compress column spacing
\begin{tabular}{cccccccc}
\hline
\toprule
\textbf{Model} & \textbf{Data} & \textbf{Succ.} & \textbf{TL} & \textbf{NE↓} & \textbf{SR↑} & \textbf{OSR↑} & \textbf{SPL↑} \\ 
\midrule
\hline
\textbf{GPT-4o mini}
    & Eng        & \textbf{100} & 17.6 & \textbf{6.98} & \textbf{21.0} & \textbf{46.0} & \textbf{13.0} \\ 
    & Ar         & \textbf{100} & 17.7 & 7.18 & 20.0 & 36.0 & 9.34 \\ 
    & Mixed      & \textbf{100} & 17.1 & 7.87 & 16.0 & 37.0 & 8.08 \\ 
\hline

\textbf{Phi-3 med}
    & Eng        & 41  & 6.89 & 7.65 & 7.32 & 7.32 & 5.66 \\ 
    & Ar         & 18  & 2.36 & 8.51 & 0.00 & 0.00 & 0.00 \\ 
\hline
\textbf{Llama 3 8B}
    & Eng        & \textbf{100} & 8.21 & 8.20 & 4.00 & 8.00 & 2.73 \\ 
    & Ar         & 96  & 7.54 & 8.34 & 3.12 & 5.21 & 1.33 \\ 
\hline
\textbf{Jais 30B}
    & Eng        & 95  & \textbf{0.68} & 8.45 & 0.00 & 0.00 & 0.00 \\ 
    & Ar         & 82  & 0.78 & 8.35 & 0.00 & 0.00 & 0.00 \\ 
\hline
\bottomrule
\end{tabular}
\label{tab:performance_metrics}
\end{table}

\subsection{Qualitative Assessment}

% \begin{table}[ht]
% \caption{Qualitative Analysis of the LMs with English and Arabic Datasets}
% \centering
% \small % Set font size to 9-10pt Roman
% \setlength{\tabcolsep}{1mm} % Compress column spacing
% \begin{tabular}{p{1.8cm}p{3cm}p{3cm}} % Adjust column widths
% % \begin{tabular}{ccc} % Adjust column widths
% \hline

% \toprule
% \textbf{Model} & \textbf{Pros} & \textbf{Cons} \\
% \midrule
% \hline

% \textbf{GPT-4o mini} & Handled English and Arabic tasks successfully & Not comparable to fine-tuned methods \\
% \hline
% \textbf{Phi-3 med} & Worked well for English instructions. Generated thoughts in Arabic & Parsing errors. Prompt engineering challenges. Failed in Arabic-instructed navigation \\
% \hline
% \textbf{Llama 3 8B} & Worked well for English and Arabic instructions & Fell behind GPT-4o mini performance. Lacked reasoning capabilities \\
% \hline
% \textbf{Jais 30B} & None & Lacked reasoning capabilities. Failed to understand prompts \\
% \hline
% \bottomrule
% \end{tabular}
% \label{tab:model_comparison}
% \end{table}

Similar to the quantitative evaluation, the qualitative evaluation was performed using both Arabic and English instructions, providing insights into how the input language affected the models' behavior and reasoning processes. The LM's planning capabilities and whether it is multilingual or not were observed to have a greater impact than the input language of instruction. 

We evaluated the models on three planning specific metrics:
\begin{itemize}
    \item \textbf{Goal Decomposition Accuracy}: Measured how well the model broke down a high-level goal into actionable steps.
    \item \textbf{Sequential Consistency}: Evaluated whether the model followed a logical sequence of actions without contradicting previous decisions.
    \item \textbf{Adaptability}: Assessed the model's ability to dynamically adjust its plan when encountering unforeseen environmental changes.

\end{itemize}

\subsubsection{Successful: English Instructions and Thoughts}
\begin{figure}[h]
    \centering
    \includegraphics[width=1.0\linewidth]{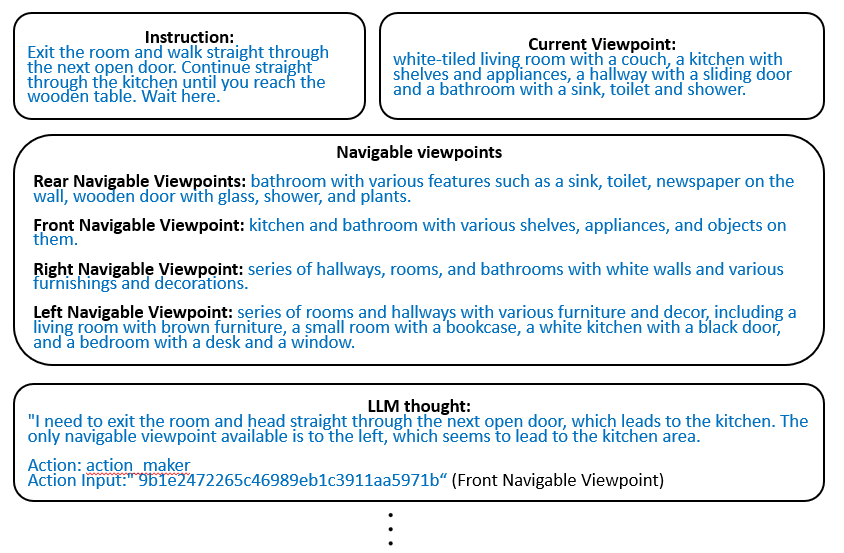}
    \caption{Successful navigation example with English instructions and thoughts}
    \label{fig:Successful}
\end{figure}

Figure \ref{fig:Successful} shows an example where the agent successfully understood the instruction and navigated to the desired area using English instructions and thoughts. The model demonstrated accurate goal decomposition and sequential consistency, effectively aligning its actions with the given instructions.

\subsubsection{Successful: Arabic Instructions and Thoughts}
\begin{figure}[h]
    \centering
    \includegraphics[width=1.0\linewidth]{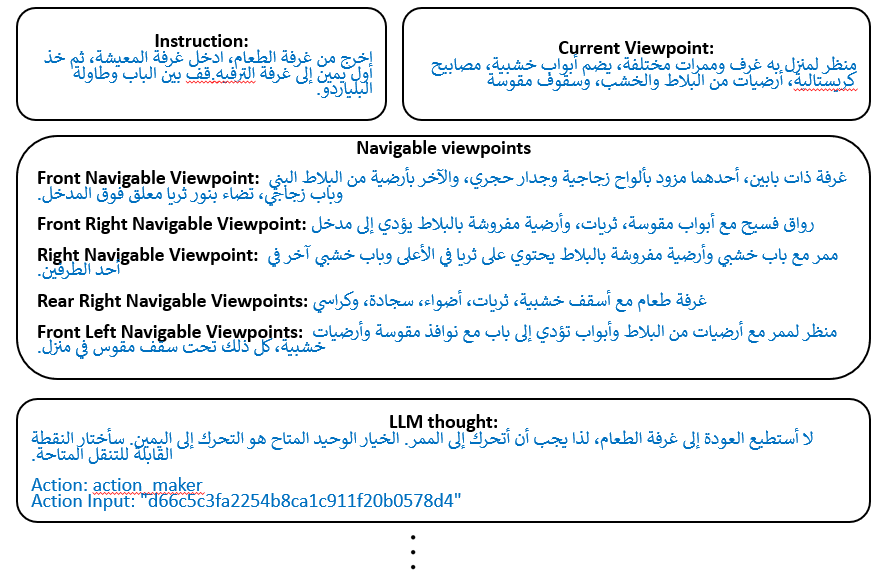}
    \caption{Successful navigation example with Arabic instructions and thoughts}
    \label{fig:Arabic}
\end{figure}

An example of our translated dataset with Arabic instructions and Arabic thoughts is provided in Figure \ref{fig:Arabic}. The agent successfully demonstrated planning and reasoning capabilities in Arabic. Despite the additional complexity of processing instructions in a non-English language, the model maintained goal decomposition accuracy and logical planning.

\subsubsection{Parsing Error}
\begin{figure}[h]
    \centering
    \includegraphics[width=1.0\linewidth]{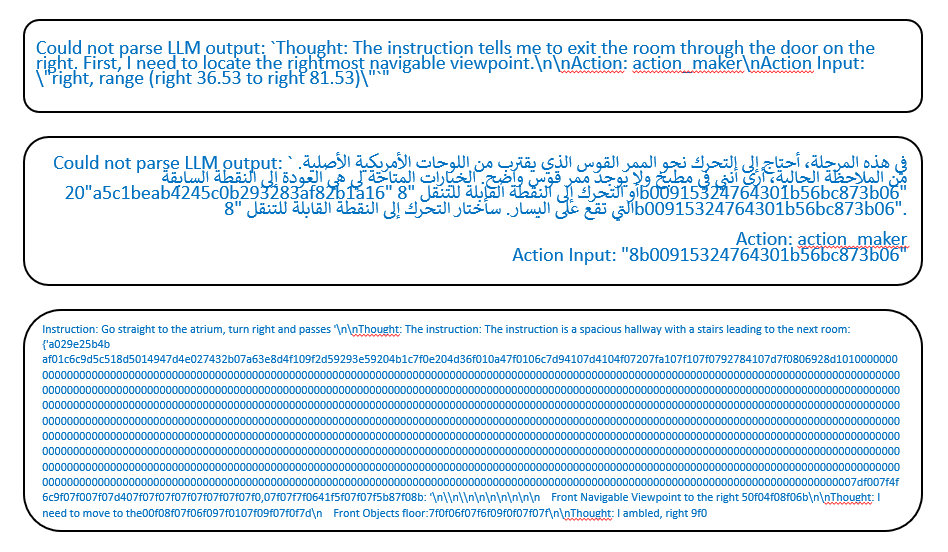}
    \caption{Example of parsing errors}
    \label{fig:Errors}
\end{figure}

As shown in Figure \ref{fig:Errors}, parsing errors occur when the model fails to output the thought and action in the expected format. These errors include incorrect formatting, missing action outputs, or hallucinating non-existent viewpoints. Such issues directly affect the sequential consistency and adaptability of the model, highlighting areas for improvement in prompt design and system parsing mechanisms.

\subsubsection{Jais Failing}
\begin{figure}[h]
    \centering
    \includegraphics[width=1.0\linewidth]{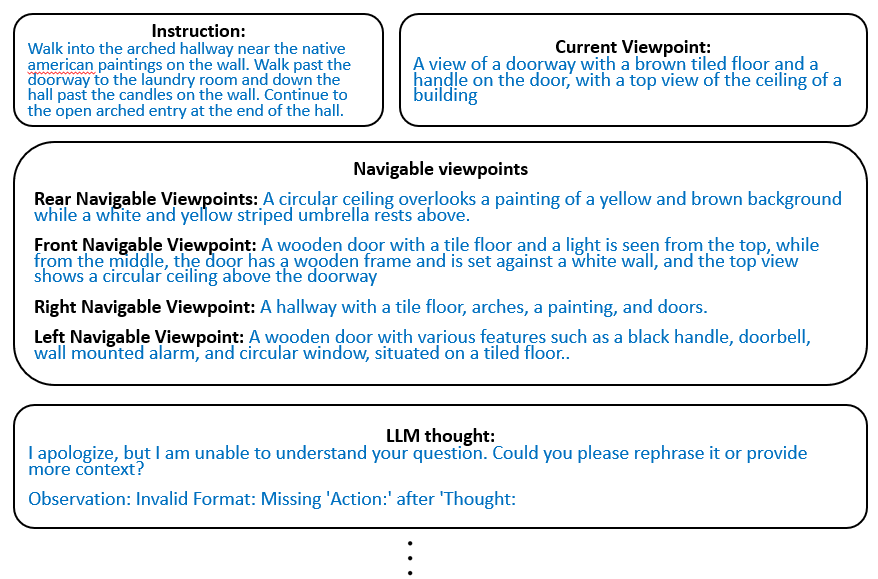}
    \caption{Example of Jais failing}
    \label{fig:Jais_Error}
\end{figure}

Jais often fails as shown in Figure \ref{fig:Jais_Error}. These failures indicate an inability to perform navigation tasks or comprehend the question. This behavior likely stems from the lack of instruction-based fine-tuning for navigation tasks and system prompt constraints, which limit its planning capabilities. Specifically, Jais struggled with sequential consistency and goal decomposition, suggesting that enhancements in task-specific fine-tuning could mitigate these issues.

By discussing planning-specific metrics and weaknesses, this assessment provides actionable insights into improving language models for navigation and task-planning scenarios.

\section{Limitations and Future Work}

\label{s:limitations}
%DESCRIPTION: Mention the limitations of proposed method and give some future directions.
The limitations of the proposed work can be summarized as follows:
\begin{itemize}
\item \textbf{Lack of Visual Features:} The visual images are not directly processed using an image encoder; instead, an image-to-text descriptor is used, resulting in information loss. As a result, we only depend on the textual depiction of visual scenes for language models.     
\item \textbf{Zero-Shot Reasoning Ability:} The language models used in this study were not fine-tuned for the specific downstream task. Instead, we relied on their zero-shot reasoning and planning capabilities for unseen tasks, which were falling behind fine-tuned models \cite{NavGPT}.

\item \textbf{Object History Tracking:} The history module summarizes previous observations into a sentence, which may result in omitting some details from earlier observations.
\item \textbf{Translated Dataset (Instructions, Observations, Objects List):} Machine translation, especially for linguistically complex languages like Arabic, is rarely perfect and can introduce errors or ambiguities. This can lead to information loss, altered context, or misrepresented semantics, which in turn affects the model's ability to generalize and make accurate predictions.

\item \textbf{Real-World Deployment:} NavGPT relies on a predefined graph of navigable viewpoints and discrete high-level actions. This contrasts with the continuous nature of real-world environments, where agents require low-level actions (for example,  move forward 0.25 meters or turn 30 degrees). However, NavGPT mitigates the sim-to-real gap by using a language model for planning and visual text descriptors for perception.

\end{itemize}

Future work can address these limitations by incorporating a dedicated vision encoder to directly process visual features, avoiding information loss caused by text-only descriptions. In addition, data sets can be translated into Arabic by human annotators to improve the quality and accuracy of instructions. Exploring state-of-the-art Arabic-centric models like SILMA \cite{silma} and ALLaM \cite{allam} offers the potential to improve Arabic language support. Finally, training and fine-tuning models on task-specific training data instead of relying on zero-shot predictions is another promising direction.

\section{Conclusion}
\label{s:conclusion}
%DESCRIPTION: Conclude/summarize your overall project in one paragraph.
This work explored the impact of language on VLN tasks by comparing multilingual SLMs with the Arabic-focused LLM, Jais, in processing and planning navigation instructions in both English and Arabic. We augmented the R2R dataset with machine-translated Arabic instructions and evaluated performance within the NavGPT framework. The results revealed that the robustness of a model's reasoning and planning capabilities mattered more than the language itself. GPT-4o mini demonstrated strong performance in both languages, outperforming Llama 3 8B by five times. However, other models, such as Phi-3 medium and Jais, struggled due to parsing issues and limited reasoning and planning capabilities, with Jais scoring 0 SR in both languages. Phi-3's poor performance with Arabic was attributed to its non-multilingual nature. These findings emphasize the need for robust, multilingual models to improve autonomous systems in Arabic-speaking regions, where language-specific models may be insufficient.

% \texttt{\textbackslash appendix}
% \texttt{\textbackslash section\{Heading\}} 
% \input{Appendix}

% \emph{Ethical Statement}
 
\section*{Acknowledgements}
We would like to express our gratitude to Mohamed Bin Zayed University of Artificial Intelligence (MBZUAI) for providing the resources and support necessary to complete this research. We are especially thankful to Professor Ian Reid and Dr. Dong An for their invaluable guidance and expertise throughout this project. Additionally, we extend our appreciation to Dr. Abdulrahman Mahmoud for his constructive feedback and support, as well as Sarim Hashmi, our teaching assistant for the AI project course, for his insightful discussions and assistance during the development of this work.

\clearpage
\bibliography{custom}

\end{document}